\documentclass[10pt]{article} 

\usepackage[accepted]{rlj} 

\usepackage{amssymb}            
\usepackage{mathtools}          
\usepackage{mathrsfs}           
\usepackage{graphicx}           
\usepackage{subfigure}
\usepackage[space]{grffile}     
\usepackage{url}                
\usepackage{lipsum}             

\usepackage{multicol}
\usepackage{algorithm}
\usepackage{algpseudocode}
\usepackage{caption}
\usepackage{wrapfig}

\definecolor{vicBlue}{RGB}{73,143,193}
\definecolor{diaynGreen}{RGB}{63,162,145}
\definecolor{lsdPurple}{RGB}{95,86,159}

\usepackage{amsthm}
\theoremstyle{definition}
\newtheorem{definition}{Definition}

\title{Focused Skill Discovery: Learning to Control Specific State Variables while Minimizing Side Effects}

\setrunningtitle{Focused Skill Discovery}

\author{Jonathan Cola\c{c}o Carr\textsuperscript{1,*,$\dagger$}, Qinyi Sun\textsuperscript{2,$\dagger$}, Cameron Allen\textsuperscript{3,*}}

\emails{jonathan.colacocarr@mail.mcgill.ca, wendysun@mit.edu, camallen@berkeley.edu}

\affiliations{
$^{1}$\textbf{McGill University and Mila - Quebec AI Institute}\\
$^{2}$\textbf{Massachusetts Institute of Technology}\\
$^{3}$\textbf{University of California, Berkeley}\\
\par 
$^*$ Corresponding Authors\\
$^\dagger$ Work done while at the Center for Human-Compatible AI
}

\contribution{
    This paper presents a general method for modifying existing skill discovery algorithms such that the learned skills control individual state variables.
    }
    {
    Prior work by~\citet{hu2024dusdi} explores a similar method, but is limited to skills that are discovered by maximizing mutual information. Our method is compatible with a wider variety of skill discovery methods, which we demonstrate with Lipschitz-constrained Skill Discovery (LSD)~\citep{park2022lsd}, as well as with two mutual-information driven skill discovery methods~\citep{gregor2017vic,eysenbach2018diayn}.
    }
\contribution{
    We show that applying our algorithm to three skill discovery algorithms improves exploration efficiency by a factor of three as compared to their original variations.
    }
    {
    We assess exploration efficiency by measuring the fraction of unique final states that can be reached after executing all possible skill-chain combinations of a given length, and then calculating the area under the curve for different skill-chain lengths.
    }

\contribution{
    We show that our approach can learn skills that automatically avoid unwanted side effects when the goal is underspecified (i.e. when the goal is only specified explicitly for a subset of the state variables, but changes to the others should still be minimized).
    }
    {
    This is an important area of research for improving the safety and effectiveness of learned skills. Prior work by~\citet{turner2020conservative} and~\citet{krakovna2020avoiding} has discussed underspecified objectives but has not explored how skill discovery as a pre-training step can help mitigate these effects.
    }
    
\contribution{
    Compared to an existing method, we show that our approach learns skills that can be significantly more effective in downstream tasks where side effects should be minimized.
    }
    {
    We measure the effectiveness in downstream tasks using the number of steps to reach the goal vs. training episode. The method we compare against~\citep{hu2024dusdi} aims to learn skills that control individual state variables while minimizing side effects, but is less effective than our method on two tasks, both in terms of final performance and learning efficiency.
    }

\summary{
Skills are essential for unlocking higher levels of problem solving. A common approach to discovering these skills is to learn ones that reliably reach different states, thus empowering the agent to control its environment. However, existing skill discovery algorithms often overlook the natural state variables present in many reinforcement learning problems, meaning that the discovered skills lack control of specific state variables. This can significantly hamper exploration efficiency, make skills more challenging to learn with, and lead to negative side effects in downstream tasks when the goal is under-specified. We introduce a general method that enables these skill discovery algorithms to learn \textit{focused skills}---skills that target and control specific state variables. Our approach improves state space coverage by a factor of three, unlocks new learning capabilities, and automatically avoids negative side effects in downstream tasks.
}
\newcommand{\vcomp}{\mathcal{V}_z^{\,\mathsf{c}}}
\definecolor{pale-yellow}{HTML}{FFF9E6}
\definecolor{soft-peach}{HTML}{FFF1E6}
\definecolor{light-lilac}{HTML}{F5F0FF}
\definecolor{mint-cream}{HTML}{E8FFF5}
\definecolor{sky-tint}{HTML}{E9F7FF}

\newtcolorbox{examplebox}{%
    colback=soft-peach!80,  
    colframe=white,             
    arc=2mm,                    
    outer arc=2mm,              
    boxrule=0pt,                
    width=\textwidth,           
    boxsep=4pt,                 
    left=5pt,                   
    right=5pt,                  
    top=5pt,
    bottom=5pt
}

\begin{document}

\makeCover  
\maketitle  

\begin{abstract}
  
Skills are essential for unlocking higher levels of problem solving. A common approach to discovering these skills is to learn ones that reliably reach different states, thus empowering the agent to control its environment. However, existing skill discovery algorithms often overlook the natural state variables present in many reinforcement learning problems, meaning that the discovered skills lack control of specific state variables. This can significantly hamper exploration efficiency, make skills more challenging to learn with, and lead to negative side effects in downstream tasks when the goal is under-specified. We introduce a general method that enables these skill discovery algorithms to learn \textit{focused skills}---skills that target and control specific state variables. Our approach improves state space coverage by a factor of three, unlocks new learning capabilities, and automatically avoids negative side effects in downstream tasks.  
\end{abstract}

\section{Introduction}

Skills are learned behaviours that allow an agent to decompose a challenging problem into a set of easier sub-problems.
In reinforcement learning (RL), a key challenge is \textit{skill discovery}: finding a useful collection of skills, either from the agent’s experiences or from an explicit task description.
The main difficulty stems from needing to determine the best way to decompose a given problem before the agent has been told which problem to solve.

A popular approach to discovering skills is to find ones that can reliably reach different areas of the state space, allowing the agent to both control and explore its environment.
Skills learned in this way facilitate control, since each skill will consistently bring the agent to the same area of the state space.
They improve exploration, since the set of states visited by each skill is unique.
These skills can be discovered without any task description by maximizing the mutual information between the agent’s selected skill and the state of the environment, or by aligning the representations of skills and states in a latent space.


However, while this approach is effective at generating a diverse set of behaviours, those behaviours tend not to provide the agent with much actual control over the individual state variables present in many reinforcement learning environments.
For example, in a robot navigation task where the state is decomposed into the positions of the agent and other objects, skills might learn to navigate to different positions without learning to collect---or avoid--- specific objects. 
The issue with these objectives is that skills are only encouraged to reach different states, regardless of the individual state variables that are changed in the process.

Learning a collection of \textit{focused skills} that change one state variable at a time offers a number of benefits. First, it can significantly improve exploration when combining skills. For example, if one skill picks up a wrench while another skill picks up a hammer, then the agent can pick up both the tool and the hammer by executing one skill after the other. Second, these skills avoid making unnecessary changes to the environment, which make learning more efficient on downstream tasks where some state variables don't need to change. Lastly, by avoiding changes to non-target state variables, focused skills significantly limit the agent's tendency to pursue alternative objectives when the reward function is misspecified.

%

\paragraph{Our Contributions.} We introduce a general method that enables skill discovery algorithms to control specific state variables. 
Rather than treating the agent's state as a unified whole, we leverage the given factored state representation to learn skills focused on changing just one state variable at a time. 
We apply our method to a variety of skill discovery algorithms and show that focused skills are able to reach three times as many states for the same number of skill executions compared to their unfocused counterparts.
In downstream task, we show that these skills can solve problems that their un-focused skills struggle with and can automatically avoid side effects with no modification to the agent's goal. 
We compare our method to a recent skill-focusing method of~\citet{hu2024dusdi}, highlighting that our method is more effective at learning focused skills and achieves stronger performance on downstream tasks.

\section{Background}

We model interactions between an agent and its environment as a factored Markov decision process.

\begin{definition}
    A \textbf{factored Markov decision process} $(\mathcal{S}, \mathcal{A}, P, R,\mu, \gamma)$ consists of a set of factored states $\mathcal{S}=\mathcal{S}^1\times\cdots\times \mathcal{S}^N$, a set of actions $\mathcal{A}$, a transition probability function $P: \mathcal{S}\times \mathcal{A}\times \mathcal{S}\to [0,1]$, a reward function $R:\mathcal{S}\times \mathcal{A}\to \mathbb{R}$, an initial state distribution $\mu$ and a discount factor $\gamma\in [0,1)$. 
\end{definition}

Many existing RL domains are factored Markov decision processes (MDPs). For example, the state of a cart-pole balancing task might be specified in terms of the positions of the cart and pole, along with their respective linear and angular velocities. The state of a board game task might be specified as a list of locations and piece types. In this paper, we assume that each state variable $s^i$ is in $\mathbb{R}^{d_i}$.


We use capital $R$ to denote the reward of the MDP, which is not observed when learning skills. 
We assume learning proceeds in two phases: first, the agent interacts with the environment reward-free, and has the opportunity to construct a set of skills (defined below). Next, the agent must learn a policy over its actions and skills that maximizes expected return, the discounted sum of future rewards from $R$. 

We model skills using the options framework~\citep{SUTTON1999181}, which we modify slightly to allow skills to depend on their interaction history. A \textit{history} $h_t=(s_0,a_0,\ldots,a_{t-1},s_t)$ is a sequence of state-action pairs that begins and ends with states. $\mathcal{H}$ denotes the set of all histories of finite length. For each state variable $i$, we also let $\mathcal{H}^i$ be the set of finite sequences of the form $(s_0^i,a_0,\ldots,a_{t-1},s_t^i)$.

\begin{definition}
    A \textbf{skill} $(\mathcal{I},\pi,\beta)$ consists of an initiation set $\mathcal{I}\subseteq \mathcal{S}$, a policy $\pi:\mathcal{H}\times \mathcal{A}\to [0,1]$ and a termination condition $\beta:\mathcal{H}\to [0,1]$.
\end{definition}
A skill might learn to pickup a tool, reach certain positions in a board game or swing a tennis racket. Defining skills in terms of histories rather than environment states allows us to conveniently model skills which end after a fixed number of timesteps. We assume that all skills terminate with probability one, and let $\text{Pr}_{\pi,P}(\cdot|s)$ the distribution over states induced by following policy $\pi$ from state $s$ according to the transition probability function $P$ until the skill terminates.

\begin{definition}
    A \textbf{focused skill} $(\mathcal{I},\pi,\beta,\mathcal{V})$ is a skill $(\mathcal{I},\pi,\beta)$ together with a non-empty set of \textbf{target variables} $\mathcal{V}\subset \{1,\ldots,N\}$. 
\end{definition}

The target variables $\mathcal{V}$ are a proper subset of $\{1,\ldots,N\}$; otherwise the skill would not focus on any specific variables. Changes to non-target variables are called side effects.

\begin{definition}
    The \textbf{side effects} of a focused skill $(\mathcal{I},\pi,\beta,\mathcal{V})$ in state $s_0\in\mathcal{I}$ is the expected number of state variables $j\not\in\mathcal{V}$ that differ between $s_0$ and the final state $s_T$ of the skill:
\begin{equation}
    \label{eq:side-effects}
    \sum_{s_T\in \mathcal{S}}\text{Pr}_{\pi,P}(s_T|s_0)\sum_{j\not\in \mathcal{V}}{\boldsymbol{1}}{\{s_T^j\neq s_0^j\}}
.\end{equation}
\end{definition}
A focused skill might learn to pick up a tool \textit{without} knocking over other objects, and would be said to have side effects if it knocks over objects while picking up the tool. We aim to minimize side effects with a side effects penalty.

\begin{definition}\label{def:side} Let $\mathcal{P}[N]$ be the power set of $\{1,\ldots,N\}$. A function $\ell:\mathcal{H}\times \mathcal{P}[N]\to \mathbb{R}_{\ge 0}$ is called a \textbf{side effects penalty} if, for any history $h_t\in \mathcal{H}$ and  $\mathcal{L}\in \mathcal{P}[N]$, $\ell$ satisfies the following two properties:
\begin{itemize}
    \item If $h_0^j\neq h_t^j$ for some $j\in \mathcal{L}$, then $\ell(h,\mathcal{L})>0$,
    \item If $h_0^j=h_t^j$ for all $j\in \mathcal{L}$ then $\ell(h,\mathcal{L})=0$.
\end{itemize}
\end{definition}
When the side effects penalty depends only on the initial state $s_0$ and the final state $s_t$ of a history $h_t$, we will overload notation and write $\ell(s_0,s_t,\mathcal{L})$ instead of $\ell(h_t,\mathcal{L})$.


Skill discovery is the process of finding a set of skills $\{(\mathcal{I}_z,\pi_z,\beta_z):z\in\mathcal{Z}\}$, where $\mathcal{Z}$ is an index set used to label each skill. We will often refer to the index $z$ as the skill \textit{itself}, to avoid re-writing $(\mathcal{I}_z,\pi_z,\beta_z)$.

When skills are focused, it is convenient to decompose the skill index set as $\mathcal{Z}=\mathcal{Z}^1\times\cdots\times \mathcal{Z}^N$.
The elements $z^i$, referred to as \textit{skill components}, indicate the effect that a skill $z$ has on variable $i$.
We can think of focused skills as skills $z=(z^1,\ldots z^n)$ for which $z^i\neq 0$ if and only if $i$ is a target variable of $z$.
We would expect two focused skills $z_1$ and $z_2$ to have the same effect on a target variable $i$ if $z_1^i=z_2^i$.

The skill discovery methods we consider here will all assume that the initiation set is equal to $\mathcal{S}$ and that the stopping conditions are fixed, either by terminating after a fixed number of timesteps or by adding a special ``terminate'' action to the set of actions. In this case, the goal of skill discovery is to learn a set of skill policies $\{\pi_z:z\in\mathcal{Z}\}$. Without access to the MDP reward $R$, a general strategy for skill discovery is to find a set of skill policies that maximize a skill reward.


\begin{definition}
    A \textbf{skill reward} is a function $r:\mathcal{Z}\times\mathcal{H}\to \mathbb{R}$. 
\end{definition}
An example skill reward might incentivize one skill for making coffee and another skill for making toast. When the skill reward is a function only the initial state $s_0$ and final state $s_t$ of a history $h_t$, we will overload notation and write $r(s_0,z,s_t)$ instead of $r(z,h_t)$. Skill rewards will always be denoted a with lower case $r$ and are agnostic to the MDP reward $R$.



\section{Related Work}
\label{sec:rw}
Unsupervised skill discovery aims to learn a collection of useful skills without an explicit task description. It can be thought of as a pretraining step to overcome challenges of exploration and data efficiency on downstream tasks. 

\paragraph{Mutual-information-based skill discovery.} A common approach to skill discovery is to maximize the mutual information be
tween skills and states, allowing the agent to control its environment. 
\citet{gregor2017vic}'s Variational Intrinsic Control (VIC) algorithm maximizes the conditional mutual information $I(Z;S_T\vert s_0)$, where $s_0$ is a starting state, $Z$ is sampled from a distribution $\nu$ over $\mathcal{Z}$ and $S_T$ is sampled by following skill $Z$ from start state $s_0$ until termination. Since this mutual information is challenging to compute, VIC maximizes the lower bound developed by~\cite{agakov2004im}:
\begin{align*}
I(Z;S_T\vert s_0) &= -H(Z|S_T, s_0) + H(Z |s_0)\\
&= \mathbb{E} [\log p(Z|S_T,s_0)] - \mathbb{E}[\log \nu(Z|s_0)]\\
&\ge \mathbb{E}[\log d(Z|S_T,s_0)]  - \mathbb{E}[\log \nu(Z|s_0)]
.\end{align*}
Here $H(\cdot | \cdot)$ is the conditional entropy. The \textit{skill discriminator} $d(Z|S_T,s_0)$ is a learned estimate of $p(Z\vert S_T,s_0)$, the posterior distribution over skills. This lower bound gets tighter as $d$ approaches the posterior distribution over skills. Therefore, the goal of VIC is to learn \textit{both} the skill policies \textit{and} a good skill discriminator. To achieve this, \citeauthor{gregor2017vic} train a set of skill policies to maximize the reward
\begin{equation}
\label{eq:vic-reward}
    r_{\text{VIC}}(s_0, z, s_T) = \log (d(z|s_T,s_0)) - \log \nu(z|s_0)
,\end{equation}
while updating $d$ to estimate the posterior distribution over skills. In a similar spirit, \citeauthor{eysenbach2018diayn}'s A Diversity Is All You Need (DIAYN) algorithm maximizes the mutual information between skills and all states that the skill visits during its execution, discovering skills with the reward
\begin{equation}
\label{eq:diayn-reward}
    r_{\text{DIAYN}}(s_0, z, s_t) = \log (d(z|s_t)) - \log \nu(z)
.\end{equation}
A wide range of innovations to this basic approach have been studied in the literature~\citep{achiam2018variational,sharma2020dynamics,campos2020explore,hansen2020Fast,zhang2021hierarchical,kim2021ibol,liu2021aps}. While maximizing mutual information produces diverse skills,~\citet{park2022lsd} observe that these approaches struggle to discover skills for covering long distances in the state space.

\paragraph{Lipschitz-based skill discovery.} Recent work has explored Lipschitz constraints to discover skills which maximize the distance traveled during skill execution. Lipschitz-constrained Skill Discovery (LSD)~\citep{park2022lsd} learns a Lipschitz function $\phi:\mathcal{S}\to\mathcal{Z}$ and a set of skills that maximize the reward
\begin{equation}
\label{eq:lsd-reward}
r_{\text{LSD}}(s_0,z,s_T) = \langle\phi(s_T)-\phi(s_0),z\rangle
,\end{equation}
where $\langle\cdot,\cdot\rangle$ is the inner product. This reward encourages $\phi(s_T)-\phi(s_0)$ to be large in the direction of the skill $z$. Due to the Lipschitz constraint ($\|\phi(s')-\phi(s)\|\le \|s'-s\|$ for all $s,s'\in \mathcal{S}$), maximizing $\phi(s_T)-\phi(s_0)$ requires the Euclidean distance $\|s_T-s_0\|$ to be large. Lipschitz-based skill discovery algorithms can also learn skills that maximize non-Euclidean notions of distance~\citep{park2023controllability} and have recently achieved state-of-the-art results in pixel-based environments~\citep{park2024metra}.


\paragraph{Leveraging State Variables in Skill Discovery.} Several methods have used state variables to improve skill discovery. Skills can be encouraged to achieve subgoals on specific variables~\citep{lee2020Learning,choi2023unsupervised} or to cause specific interactions between state variables~\citep{hu2022causality,wang2024skild}. While these algorithms improve control of state variables, they do not penalize side effects to other state variables.
%
%
The benefits of skills which minimize side effects has been studied in the planning literature~\citet{allen2021focused_macros}.

The most relevant point of comparison to our work is Disentangled Unsupervised Skill Discovery (DUSDi)~\citep{hu2024dusdi}, a method designed to make mutual-information-based skill discovery algorithms learn focused skills.
DUSDi maximizes the mutual information between skills and values on target variables while minimizing the mutual information between skills and values of non-target variables. However, we show that this objective is not always effective at learning focused skills because the mutual information penalty does not explicitly minimize side effects. We provide a more detailed comparison with DUSDi in Section~\ref{subsec:dusdi-comparison} and compare the two methods in our experiments. A second distinction between DUSDi and our approach is that our approach is compatible with non mutual-information-based skill discovery algorithms, such as Lipschitz-based algorithms.

\section{Focused Skill Discovery}
\label{sec:focused-skill-discovery}


\begin{algorithm}[htpb]
\caption{Focused Variational Intrinsic Control}\label{alg:fvic}
\begin{algorithmic}
\For{episode $=1,M$}
\State Sample $s_0$ from the initial state distribution $\mu$
\State Sample skill $z$ from $\nu(\cdot | s_0)$
\State Follow policy $\pi_z$ until termination state $s_T$
\For{$i$ in $\mathcal{V}_z$}
\State Update the skill discriminator $d_{i}$ from $(s_0^i, z^i, s_T^{i})$ 
\EndFor
\State Calculate the reward $r_{\text{focused-VIC}}(s_0, z, s_{T})$ using Equation~\ref{eq:focused-vic}.
\State Update $\pi_z$ to maximize $r_{\text{focused-VIC}}$
\State Update option prior $\nu(\cdot | s_0)$ based on $r_{\text{focused-VIC}}$
\EndFor
\end{algorithmic}
\end{algorithm}
In this section, we present \textit{focused skill discovery}: a general method of transforming existing skill discovery algorithms into ones that discover focused skills. Our method can be applied to any skill discovery algorithm that learns skills in a factored Markov decision process using a skill reward. We will first describe our concept in general terms and then show how it can be applied to mutual-information-based and Lipschitz-based skill discovery algorithms. Lastly, we describe the key differences between our method and DUSDi~(\citeyear{hu2024dusdi}) which allow our method to minimize side effects more effectively.

To create a focused skill discovery algorithm from a baseline skill discovery algorithm, we modify the baseline algorithm's skill reward. Focused skill rewards incentivize skills to control their target variables while minimizing changes to other state variables. The original skill discovery algorithm is transformed into a focused skill discovery algorithm once the baseline skill reward is replaced with the focused skill reward.

Focused skill rewards consist of two terms: one which encourages focused skills to manipulate their target variables and another which penalizes side effects. The first term is a sum of reward components $r_i:\mathcal{Z}\times\mathcal{H}^i\to\mathbb{R}$, each generated by restricting a copy of the original skill reward to $\mathcal{Z}\times\mathcal{H}^i$. The second term is a side effects penalty $\ell$, which discourages the skill from affecting non-target variables. When combined, these two terms define a focused skill reward
\begin{equation}
    r_{\text{focused}}(z,h_t) = \left[\sum_{i\in \mathcal{V}_z}r_i(z, h_t^i)\right] - \ell(h_t,\vcomp)
.\end{equation}

\begin{examplebox}
\textbf{Example 1} (Focused skill rewards for VIC and DIAYN)
To create focused skill rewards for VIC and DIAYN, we learn a separate skill discriminator $d_i$ for each target variable $i$, which predicts skills based on the values of state variable $i$. This encourages skills to reliably reach different values of their respective target variables. Starting with the VIC reward in Equation~\ref{eq:vic-reward}, we obtain the focused VIC reward
\begin{align}
    r_{\text{focused-VIC}}(s_0, z, s_{T}) &= \left[\sum_{i\in \mathcal{V}_z}\log(d_{i}(z| s_{T}^{i}, s_{0}^{i}))  - \log (\nu(z|s_0^i))\right]- \ell(s_0,s_T,\vcomp)\label{eq:focused-vic}
.\end{align}
Similarly, the DIAYN reward in Equation~\ref{eq:diayn-reward} leads to the focused DIAYN reward
\begin{align}
    r_{\text{focused-DIAYN}}(s_0, z, s_{t}) &= \left[\sum_{i\in \mathcal{V}_z}\log(d_{i}(z| s_{t}^{i}))  - \log (\nu(z))\right]-\ell(s_0,s_t,\vcomp)\label{eq:focused-diayn}
.\end{align}
\end{examplebox}
\begin{examplebox}
\textbf{Example 2} (Focused skill reward for LSD)
A focused skill reward for Lipschitz-constrained skill discovery is constructed by learning a Lipschitz-constrained function $\phi_i:\mathcal{S}^i\to\mathcal{Z}$ for each target variable $i$. From the LSD reward in Equation~\ref{eq:lsd-reward}, we derive the focused LSD reward
\begin{equation}
    \label{eq:focused-lsd}
    r_{\text{focused-LSD}}(s_0,z,s_T) = \left[\sum_{i\in\mathcal{V}_z}\langle \phi_i(s_0^i)-\phi_i(s_T^i), z\rangle\right] - \ell(s_0,s_T,\vcomp)
.\end{equation}
This reward motivates skills to maximize the distance between initial and final states on their target variables.
\end{examplebox}

When a skill reward is substituted for its focused variant, we obtain a \textit{focused skill discovery algorithm}---one which is capable of learning focused skills.
Algorithm~\ref{alg:fvic} applies this transformation to Variational Intrinsic Control. Algorithms for the focused versions of DIAYN and LSD are provided in Appendix~\ref{appendix:focused-algorithms}.

\subsection{Comparison with DUSDi}
\label{subsec:dusdi-comparison}

While both focused skill discovery and DUSDi~(\citeyear{hu2024dusdi}) aim to learn focused skills, they differ in two key ways. First, the DUSDi objective is defined through mutual information, while our method remains agnostic to how the skill reward is generated. This allows our method to be applied to a wider range of skill discovery algorithms, such as LSD (\citeyear{park2022lsd}). Second, our methods differ in how they mitigate side effects.

DUSDi mitigates side effects by minimizing the mutual information $I(Z^i,S^{\neg i})$ between skill components $Z^i$ and the values of the remaining state variables, $S^{\neg i}= (S_1,\ldots,S_{i-1},S_{i+1},\ldots S_N)$.
However, this mutual information does not necessarily penalize side effects; it only encourages skills to have the \textit{same} effects on non-target state variables. 

Having the \emph{same} effects on non-target variables does not mean that there are \emph{no} side effects. Consider the case of a robotic arm with two state variables: one for the position of the arm's gripper and another indicating whether a cup of coffee on a nearby table has been knocked over. If $i$ is the position variable of the gripper, then $I(Z^i,S^{\neg i})$ can be minimized if \textit{all} skill components for the arm gripper knock over the cup of coffee, because the components are indistinguishable from one another on the coffee cup variable if they all knock the cup over.

While this may seem like a contrived example, the issues with penalizing side effects through minimizing mutual information become apparent when state variables are \textit{entangled}---i.e. a change in one state variable is correlated with changes in other state variables. In this case, during skill discovery, the skill discriminator may begin to encourage focused skills to make unnecessary changes their non-target state variables in order to minimize mutual information. This behavior is observed empirically in our experiments.

On the other hand, focused skill discovery explicitly penalizes side effects, making it impossible to maximize a focused skill reward if skills cause unnecessary changes to their non-target variables. In our experiments below, we will study how this leads to substantial differences between skills learned with DUSDi and skills learned focused skill discovery.

\section{Experiments}
\label{subsec:skill-trajectories}
\begin{figure}[htbp]
    \centering
    \subfigure[FourRooms]{
        \includegraphics[width=0.45\textwidth]{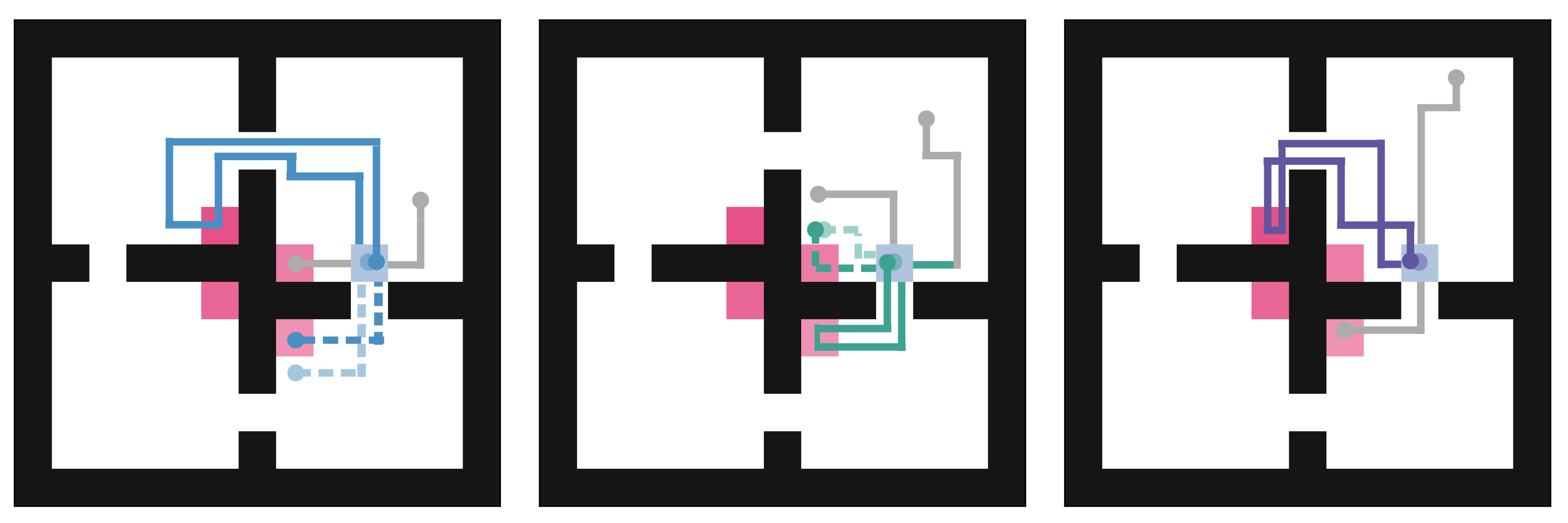}
        \label{fig:fourrooms}
    }
    \subfigure[MudWorld]{
        \includegraphics[width=0.45\textwidth]{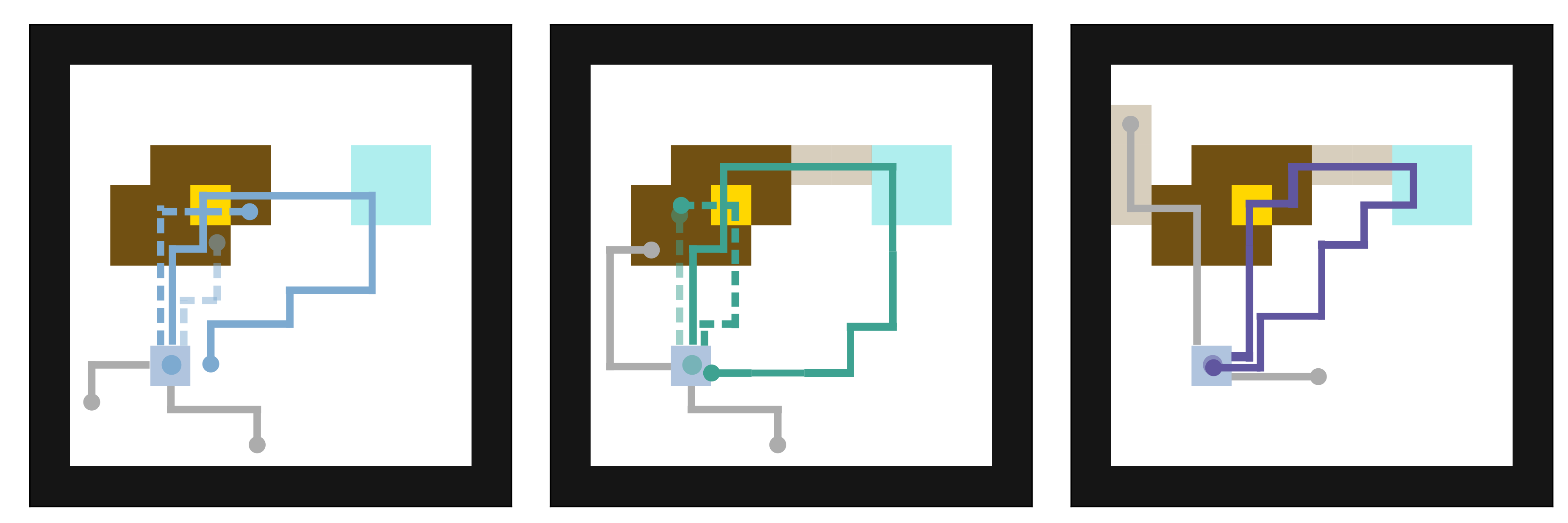}
        \label{fig:mudworld}
    }
    \subfigure[ForageWorld]{
        \includegraphics[width=0.45\textwidth]{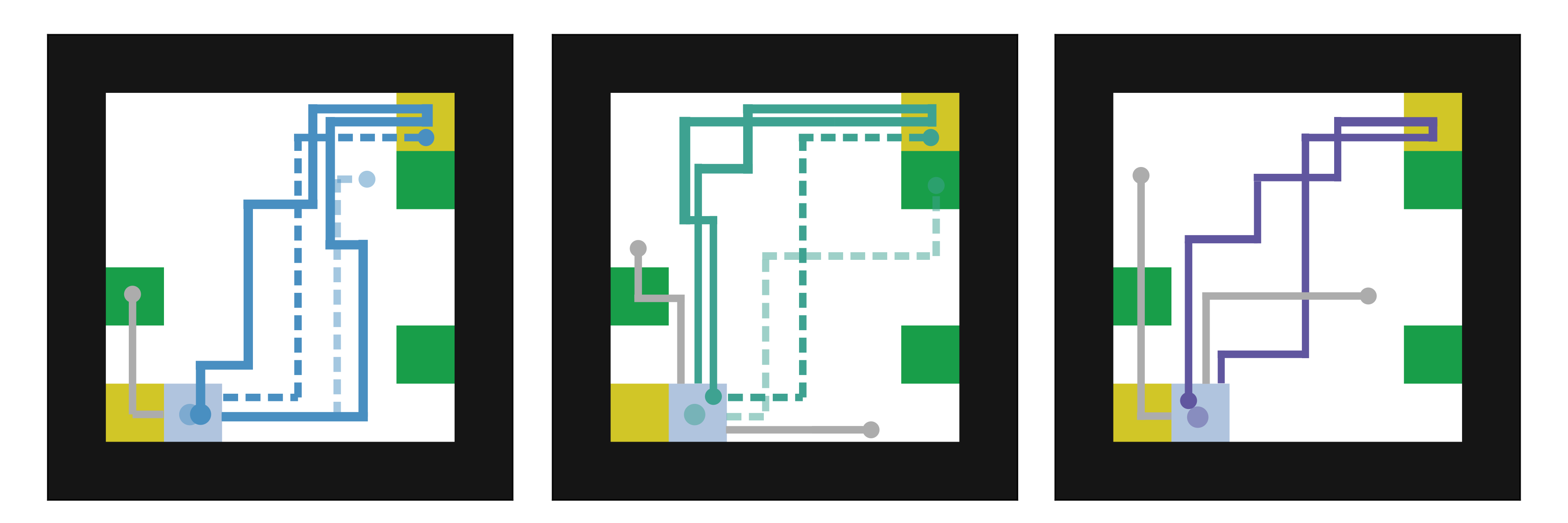}
        \label{fig:forageworld}
    }
    
    \caption{Skill trajectories for \textcolor{vicBlue}{\rule[0.5ex]{1em}{1pt}} VIC, \textcolor{diaynGreen}{\rule[0.5ex]{1em}{1pt}} DIAYN and \textcolor{lsdPurple}{\rule[0.5ex]{1em}{1pt}} LSD algorithms. Solid lines are trajectories from focused skills, dashed lines are from DUSDi skills and grey lines are from the baseline algorithms. Skills start in the blue square and terminate at the circles. Focused skills learn to collect objects and return to the start state in order to minimize side effects.}
    \label{fig:skill-traj}
\end{figure}

We seek to develop a principled understanding of how skills learned with focused skill discovery compare to baseline skill discovery algorithm and focused skills learned with DUSDi.
We consider three gridworld environments that differ in their types of interactions and their opportunities for side effects.
Qualitatively, we find that our method consistently learns skills that minimize side effects.
Focused skills dramatically outperform unfocused skills across all downstream tasks we consider, and can reach three times as many states in the same amount of skill execution steps as unfocused skills.
The difference in the side effect penalties between focused skill discovery and DUSDi leads to significantly different skills learned from the same baseline algorithms. 
When side effects must be minimized to accomplish the task, our skills outperform DUSDi, and are the only ones that can accomplish the task in our most challenging environment.
When agents optimize a proxy reward which does not penalize side effects, all focused skill discovery algorithms automatically avoid making unnecessary changes, while all other methods fall short.

We apply our method to three baseline algorithms: VIC~\citep{gregor2017vic}, DIAYN~\citep{eysenbach2018diayn} and LSD~\citep{park2022lsd}, obtaining three focused skill discovery algorithms: Focused VIC, Focused DIAYN and Focused LSD. These baselines were selected because they are have similarities with a broad range of other skill discovery algorithms, while each being unique from one another. In particular, the difference between DIAYN and VIC is largely whether the skill reward is sparse or dense, which allows us to examine the effect of sparse and dense focused skill rewards. LSD takes an entirely different approach to discovering skills. We compare these methods to DUSDi VIC and DUSDi DIAYN, the algorithms obtained by applying DUSDi to VIC and DIAYN. There is no analog of DUSDi for LSD since it only works for mutual-information-based skill discovery algorithms.


\textbf{Environments.} We learn skills in the FourRooms, ForageWorld and MudWorld environments shown in Figure~\ref{fig:skill-traj}. 
Each environment has four actions, moving the agent up, right, left, and down. When an agent selects an action to move in one direction, it moves in one of the other three directions with probability 0.1. 

FourRooms has the same map as~\citet{SUTTON1999181}, with four added tools that the agent can pick up (shown in pink). It contains five state variables: one for the agent's position and one binary variable for each tool indicating whether it has been picked up. This environment is quite challenging to explore completely: a sixteenth of the states (the ones in which all tools have been collected) are only accessible once the agent has moved through all four rooms.

ForageWorld contains yellow resources for an agent to collect, as well as three plants (green cells) that should be avoided. If the agent walks over a plant, the plant is destroyed and does not regenerate. The states in ForageWorld contain six state variables: one variable for the agent's position, two integer-valued variables for the quantity of each resource collected and three binary variables indicating which plants have been destroyed. The challenge for skills in this environment is not exploration, but how to navigate and collect resources without damaging the plants.

The MudWorld environment contains a patch of mud (brown), a piece of treasure (yellow), and a puddle (teal). The agent becomes muddy if it moves to the mud patch and, when it is muddy, it tracks mud onto the vacant (white) cells. The agent can become clean again if it moves into the puddle. We simplify the agent state in the MudWorld environment to contain four state variables: one for the agent's position, one indicating whether the agent is muddy, one variable that indicates whether the treasure has been collected and a variable for the number of muddy cells. This creates partial observability, since the agent knows only the number of muddy cells, not the location of the tracked mud. However, because the agent should always try to minimize its tracked mud, the state variables contain enough information for the agent to behave optimally. Unlike FourRooms and ForageWorld, the state variables in MudWorld are \textit{entangled}; it is impossible for the agent to collect treasure without becoming muddy. Once in the mud-patch, the only way to become less muddy is to move to the puddle, which increases the number of muddy cells.

\textbf{Implementation.} We trained all policies using tabular Q-learning with $\epsilon$-greedy exploration and a discount factor of 0.99. We trained a set of skill policies $\mathcal{Z}$ with $|\mathcal{Z}|=16$ skills in each skill discovery algorithm. 
We assign each focused skill a single target variable (i.e. $|\mathcal{V}_z|=1$). For the focused skill discovery algorithms and DUSDi algorithms, we assign two skills to target each tool, resource or treasure and use the remaining skills (8 in FourRooms, 12 in ForageWorld, 14 in MudWorld) to control the agent's position. All skills were allowed to execute for a maxmium of 40 steps in FourRooms and 20 steps in ForageWorld and MudWorld.

For the side effects penalty of focused skill rewards, we apply a weighted 2-norm $\|\cdot\|_{\boldsymbol{\lambda}}$, where $\boldsymbol{\lambda}\in \mathbb{R}^{d_1}\times \cdots \times \mathbb{R}^{d_N}$, to penalize changes on non-target variables. More precisely, for history $h_t=(s_0,a_0,\ldots,a_{t-1},s_t)$ and target variables $\mathcal{V}_z$, $\ell(h_t,\vcomp)=\|s_0-s_t\|_{\boldsymbol{\lambda}_{\mathcal{V}=0}}$, where $\boldsymbol{\lambda}_{\mathcal{V}=0}$ is the weight matrix obtained by setting the values of $\boldsymbol{\lambda}$ on target variables equal to zero, thus avoiding penalties on target variables. Using a weight matrix to control the penalty strength is useful since each state variable may have a different range of values.   For each focused skill discovery algorithm, we chose a \textit{single hyperparemeter} $\lambda>0$ and set $\lambda_{ij}$ equal to $\lambda$ divided by the maximum 2-norm between any two values on variable $j$. So, for a state variable whose values range from 0 to $k$, the weight was $\lambda / k$. This ensures that the strength of the side effect penalty for each variable is between 0 and $\lambda$. We used $\lambda=10$ for Focused VIC and Focused DIAYN and $\lambda=2$ for Focused LSD, discussing the effects for different values of $\lambda$ in Section~\ref{sec:ablations}.

Additional training details are available in Appendix~\ref{appendix:hyperparameters}.

\subsection{Qualitative Analysis of Learned Skills}

Figure~\ref{fig:skill-traj} shows skill trajectories sampled from each skill discovery algorithm.
As expected, all focused skill discovery algorithms find skills that change control individual state variables while minimizing side effects. 
The DUSDi skills are less effective at avoiding side effects; in ForageWorld and MudWorld, the skills that are designed to control the resource and treasure state variables do so while damaging nearby green cells and without cleaning off after stepping into the mud patch.
This is consistent with the differences between focused skill discovery and DUSDi described in Section~\ref{subsec:dusdi-comparison}.


\subsection{Exploration Efficiency}
\label{subsec:exploration-efficiency}
\begin{figure}
    \centering
    \includegraphics[width=\linewidth]{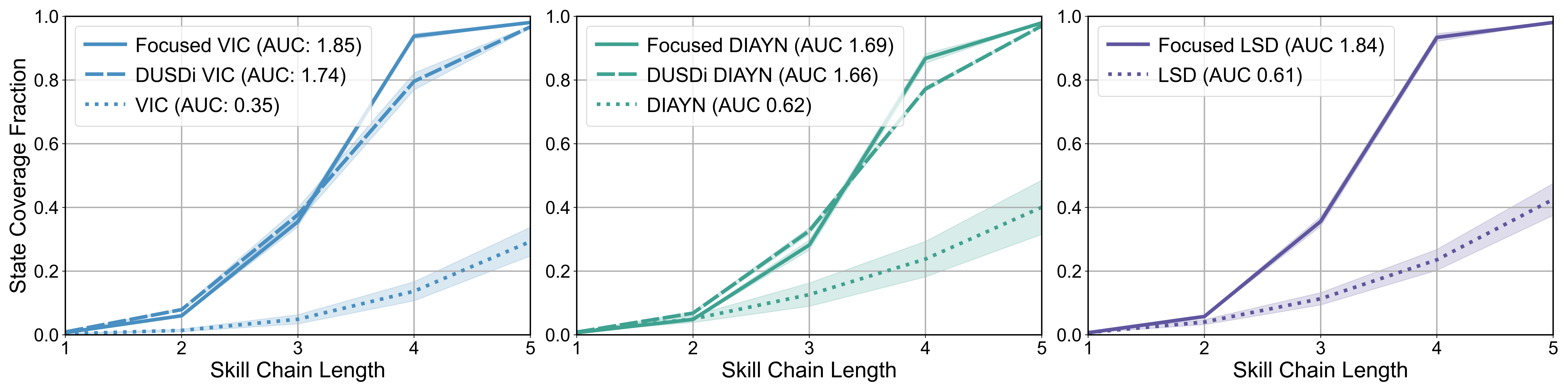}
    \caption{State coverage in the FourRooms environment. Focused skills explore three times more efficiently than unfocused skills, as measured by the Area Under the Curve (AUC).
    }
    \label{fig:expl-efficiency}
\end{figure}
One major benefit of focusing skills is that they can provide the agent with a way to structure its exploration of the environment. Equipped with skills for each tool in the FourRooms environment, agents are capable of traveling through all four rooms in just four skill execution steps. This drastically improves exploration efficiency, as measured by the Area Under the Curve (AUC) of the State Coverage Fraction vs. Skill Chain Length graph, shown in Figure~\ref{fig:expl-efficiency}. 
This measures how many states a set of skills is able to reach for a given number of skill executions.
To compute the state coverage fraction of a skill chain of length $l$, we measured the fraction of unique final states that could be reached after executing all possible skill chain combinations of length $l$ from a given start state $s_0$.
We plotted the mean state coverage fraction and 90\% confidence intervals over 10 random start states.
In all cases, the exploration benefits of focusing skills are significant: exploration improves by 5.3$\times$ for VIC (AUC 1.86 vs AUC 0.35), 2.7$\times$ for DIAYN (AUC 1.69 vs AUC 0.62) and 3.0$\times$ for LSD (AUC 1.84 vs. 0.61). This corresponds to an average improvement in exploration efficiency of 3.67$\times$ across these three methods.

The exploration efficiency of the skills learned with DUSDi and our method are comparable. This was expected, since both objectives explicitly learn skills that can pick up each tool, facilitating the agent's exploration of hard-to-reach states. Because DUSDi is limited to mutual-information-based skill discovery algorithms, there is no DUSDi variant of LSD.


\subsection{Performance on Downstream Tasks}
\label{sec:quant}
\begin{figure}
    \centering
    \includegraphics[width=\columnwidth]{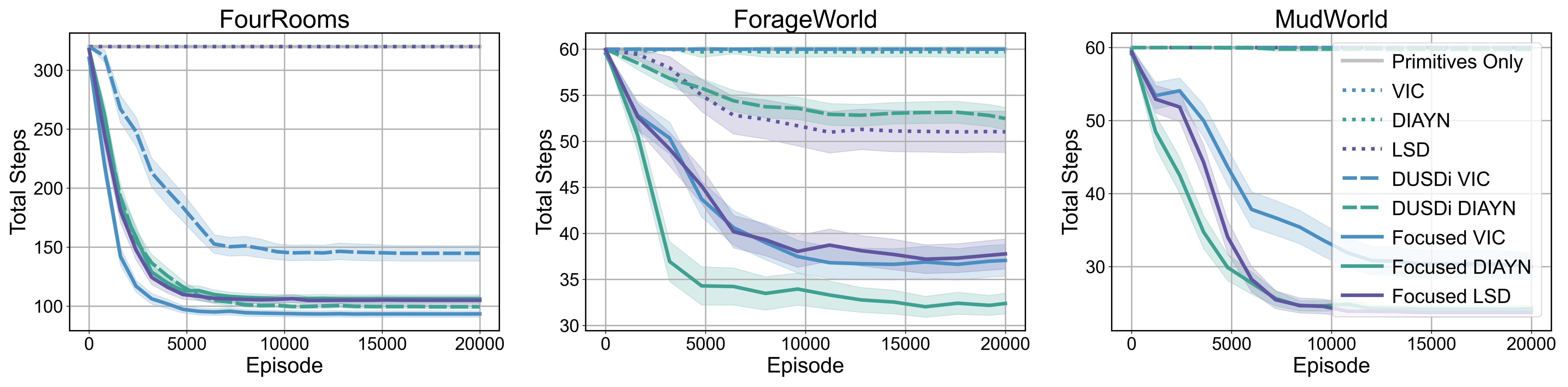}
    \caption{Learning performance in downstream tasks. Focused skills (solid lines) lead to faster learning and are the only ones which can accomplish the task in MudWorld.}
    \label{fig:efficiency}
\end{figure}

Given a set of focused skills, an agent is much more capable of solving downstream tasks.
We define downstream tasks as follows.
In all environments, the agent's goal is to collect all of the tools or resources and navigate to the bottom-right corner. In the FourRooms environment, the agent's goal is to pick up all four tools and navigate to the bottom-right corner. In the ForageWorld environment, the agent must collect two units of both resources without destroying any delicate plants. In the MudWorld environment, the agent must collect the treasure while tracking fewer than five mud cells.
In all tasks, the agent starts in the top-left corner and receives a (sparse) reward of +1 for accomplishing the task. The agent can take up to 320 steps in FourRooms, selecting at least 8 skills per episode, and up to 60 steps in ForageWorld and MudWorld, selecting at least 3 skills per episode. We conduct 50 independent training runs for each of the agents we consider, plotting the mean and 90\% confidence intervals of our results in Figure~\ref{fig:efficiency}.

Without focused skills, agents fail to accomplish their goals. Moreover, while the performance of DUSDi and focused skills are comparable in FourRooms, focused skills lead to significantly better results in ForageWorld and MudWorld, where an agent must minimize side effects in order to accomplish its goal.
This is consistent with our observations from Figure~\ref{fig:skill-traj}, where DUSDi skills seem to have the \textit{same effects }on non-target state variables, while the focused skills learn to minimize side effects.
In the MudWorld environment, only the agents with focused skills can accomplish the task.


\subsection{Underspecified Goals}
\label{subsec:underspec}
\begin{figure}[htbp]
    \centering
        \subfigure[ForageWorld]{
        \includegraphics[width=0.45\textwidth]{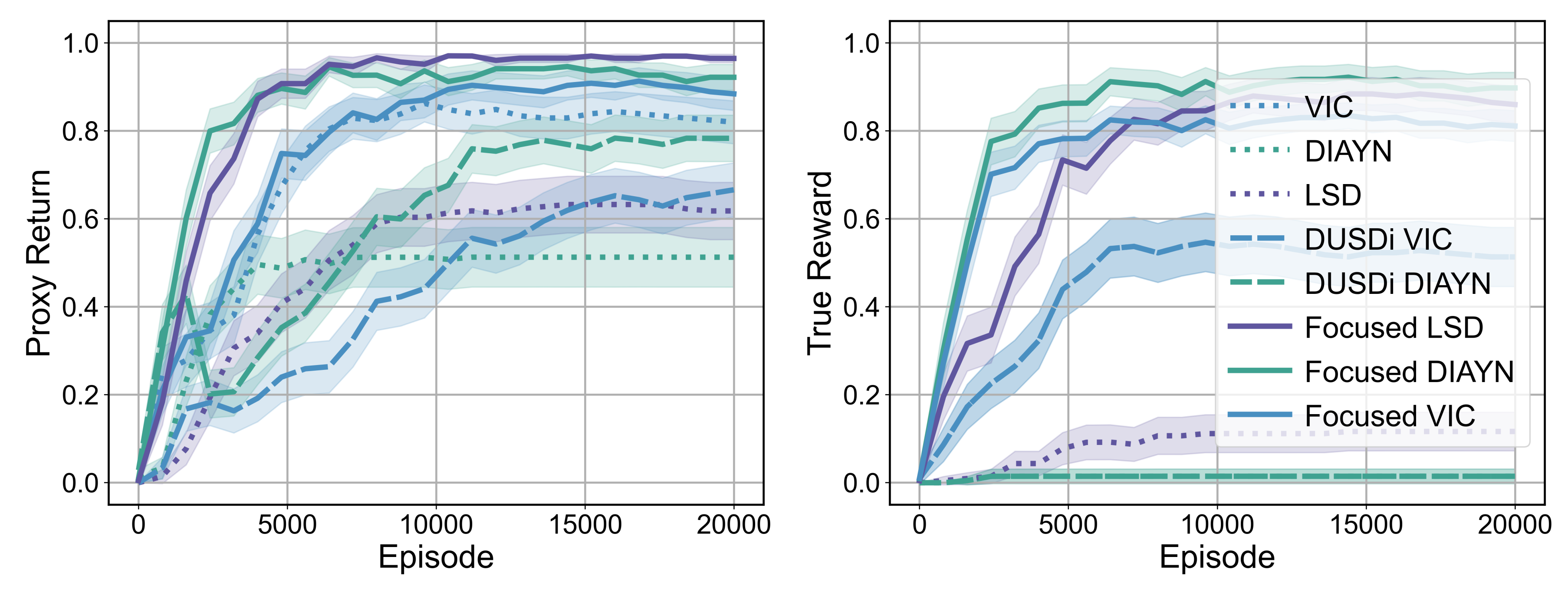}
        \label{fig:forage-underspecified-goals}
    }
    \subfigure[MudWorld]{
        \includegraphics[width=0.45\textwidth]{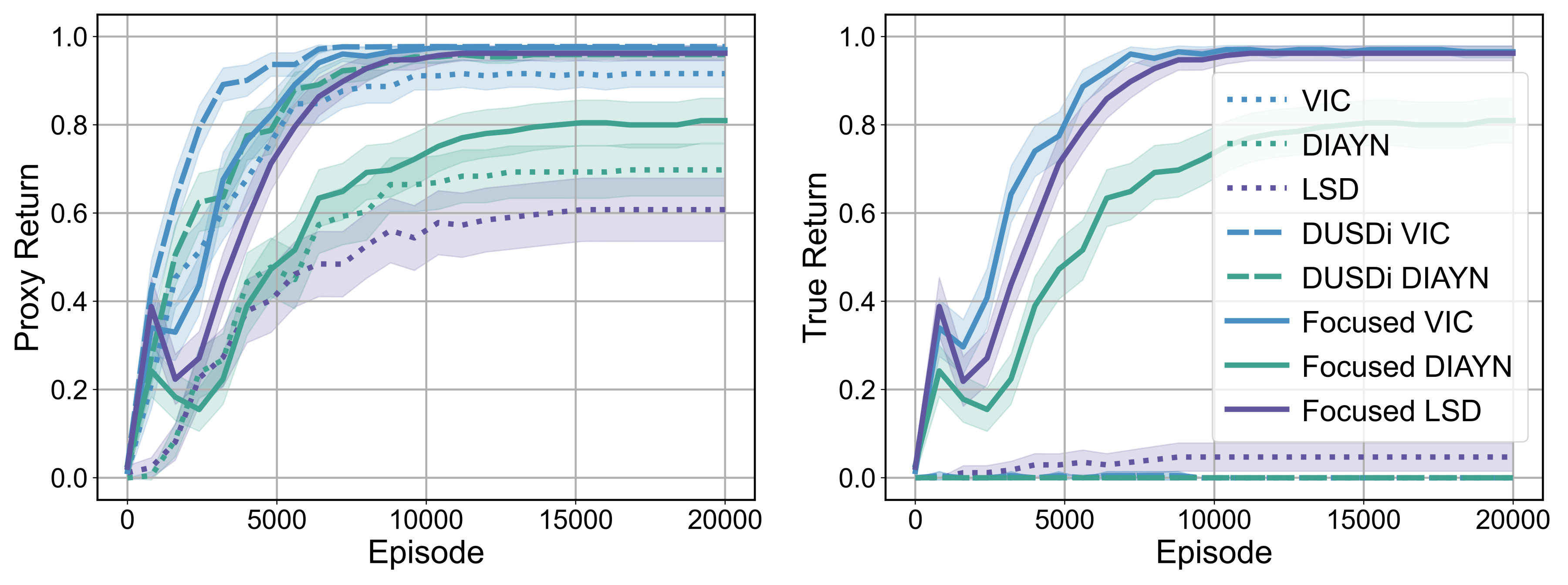}
        \label{fig:mudworld-underspecified-goals}
    }
    \caption{Task performance in ForageWorld and MudWorld when agents are trained with a proxy reward instead of the true reward. Focused skills (solid lines) are the only ones which maximize the true return when only given the proxy reward, meaning that they are the only ones that can automatically avoid making unwanted changes that are not explicit in the agent's goal.}
    \label{fig:underspecified-goals}
\end{figure}

In addition to exploration and learning benefits, focused skill discovery learns skills that automatically avoid unwanted side effects when the goals are underspecified. Figure~\ref{fig:underspecified-goals} shows the results of downstream tasks in ForageWorld and MudWorld, where the agent is not given the true task rewards from Section~\ref{sec:quant}, but instead proxy rewards which do not penalize side effects. In ForageWorld, the proxy reward gives a score of +1 for reaching the bottom-right corner after picking up all of the resources, regardless of the number of plants destroyed. In the MudWorld task, the proxy reward assigns a score of +1 for reaching the bottom-right corner after picking up the treasure and cleaning itself off, regardless of the number of tracked mud cells. Across both environments and all baseline algorithms, focused skill discovery illustrates a striking ability to accomplish the true task even if it is only given the proxy reward, meaning that it automatically avoids making unwanted changes that are not explicit in the agent's goal. In contrast, the DUSDi and un-focused skills maximize the proxy reward at the expense of the true reward.

\subsection{Ablation Studies}
\label{sec:ablations}
\begin{figure}
    \centering
    \includegraphics[width=\linewidth]{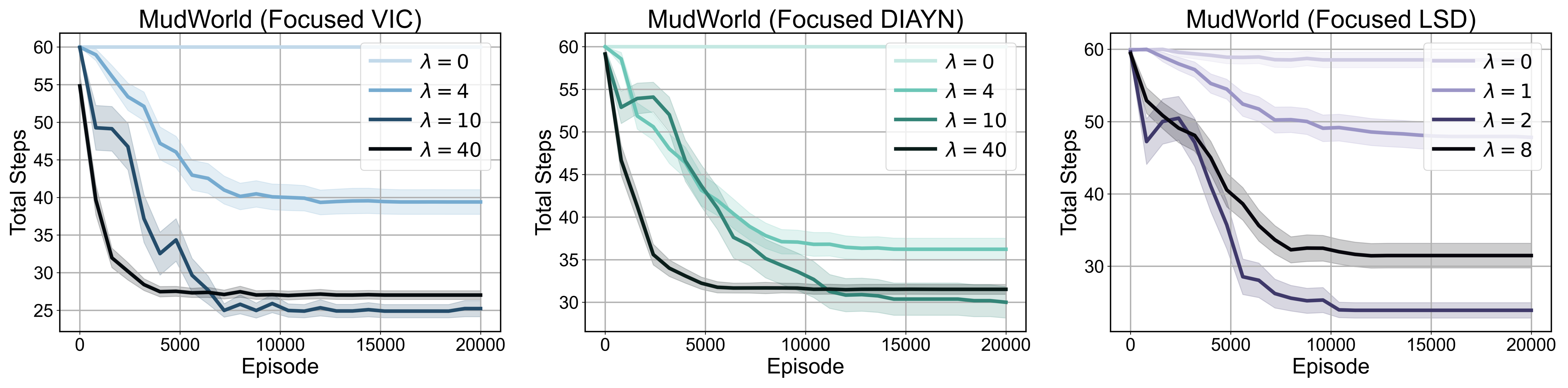}
    \caption{Impact of the side effect penalty strength on focused skills in the MudWorld domain. Skills are not effective when there is no side effects penalty (i.e. when $\lambda=0$).}
    \label{fig:ablation-penalty}
\end{figure}

The tendency for focused skills to avoid side effects in our experiments was controlled by the side effects hyperparameter $\lambda$. To better understand impact of the penalty strength, we re-ran the MudWorld task for focused skills learned with varying penalty strengths. 
We ran our experiments across 50 different training seeds, plotting the mean and 90\% confidence intervals in Figure~\ref{fig:ablation-penalty}.
In general, increasing $\lambda$ tends to improve performance, though performance degrades slightly if the penalty strength is too large. Without the side effects penalty (i.e. when $\lambda=0$), agents with focused skills are not able to accomplish the MudWorld task.



\section{Discussion \& Conclusion}
We presented focused skill discovery, a simple method that allows skill discovery algorithms to learn focused skills. Focused skills are useful not just for minimizing side effects but also for improving exploration and learning efficiency. We showed that focused skills can dramatically improve both an agent's exploration and learning capabilities while also providing a natural buffer for reward hacking when downstream rewards are under-specified. Compared to a recently proposed approach to discovering focused skills, our method showed substantial improvement, particularly in an environment where states were entangled. We are excited to keep exploring the benefits of focused skills.

There are a number of interesting avenues of future work to consider. Empirically, it would be interesting to scale these experiments up to larger environments and test our method for sets of continuous skills. Since the baseline methods we considered scale well to these settings, we are confident that focused skill discovery will also scale.
%
While pre-defined state variables play a key role in our approach, we hope to extend this idea to include other kinds of state abstractions.
There is also interesting theoretical territory to explore at the intersection of reward hacking and focused skill discovery. It seems plausible that in some cases, focused skill discovery could be guaranteed to lead to improvements an agent's capabilities while at the same time mitigating reward hacking.



\subsubsection*{Acknowledgments}
\label{sec:ack}
We would like to thank the reviewers for their thoughtful comments and suggestions. We would also like to thank Ray Luo, Raymond Chua, Padideh Nouri, Aly Lidayan, and Vivek Myers for their helpful feedback.


\bibliography{main}

\begin{thebibliography}{23}
\providecommand{\natexlab}[1]{#1}
\providecommand{\url}[1]{\texttt{#1}}
\expandafter\ifx\csname urlstyle\endcsname\relax
  \providecommand{\doi}[1]{DOI: #1}\else
  \providecommand{\doi}{DOI: \begingroup \urlstyle{rm}\Url}\fi

\bibitem[Achiam et~al.(2018)Achiam, Edwards, Amodei, and Abbeel]{achiam2018variational}
Joshua Achiam, Harrison Edwards, Dario Amodei, and Pieter Abbeel.
\newblock Variational option discovery algorithms.
\newblock \emph{arXiv preprint arXiv:1807.10299}, 2018.

\bibitem[Allen et~al.(2021)Allen, Katz, Klinger, Konidaris, Riemer, and Tesauro]{allen2021focused_macros}
Cameron Allen, Michael Katz, Tim Klinger, George Konidaris, Matthew Riemer, and Gerald Tesauro.
\newblock Efficient black-box planning using macro-actions with focused effects.
\newblock In \emph{Proceedings of the Thirtieth International Joint Conference on Artificial Intelligence}, 2021.

\bibitem[Barber \& Agakov(2004)Barber and Agakov]{agakov2004im}
David Barber and Felix Agakov.
\newblock The {IM} algorithm: a variational approach to information maximization.
\newblock In \emph{Advances in Neural Information Processing Systems}, volume~17, pp.\  201–208, 2004.

\bibitem[Campos et~al.(2020)Campos, Trott, Xiong, Socher, Giro-I-Nieto, and Torres]{campos2020explore}
Victor Campos, Alexander Trott, Caiming Xiong, Richard Socher, Xavier Giro-I-Nieto, and Jordi Torres.
\newblock Explore, discover and learn: Unsupervised discovery of state-covering skills.
\newblock In \emph{Proceedings of the 37th International Conference on Machine Learning}, volume 119, pp.\  1317--1327, 2020.

\bibitem[Choi et~al.(2023)Choi, Lee, Wang, Sohn, and Lee]{choi2023unsupervised}
Jongwook Choi, Sungtae Lee, Xinyu Wang, Sungryull Sohn, and Honglak Lee.
\newblock Unsupervised object interaction learning with counterfactual dynamics models.
\newblock In \emph{Workshop on Reincarnating Reinforcement Learning at the International Conference on Learning Representations}, 2023.

\bibitem[Eysenbach et~al.(2018)Eysenbach, Gupta, Ibarz, and Levine]{eysenbach2018diayn}
Benjamin Eysenbach, Abhishek Gupta, Julian Ibarz, and Sergey Levine.
\newblock Diversity is all you need: learning skills without a reward function.
\newblock In \emph{International Conference on Learning Representations}, 2018.

\bibitem[Gregor et~al.(2017)Gregor, Rezende, and Wierstra]{gregor2017vic}
Karol Gregor, Danilo~Jimenez Rezende, and Daan Wierstra.
\newblock Variational intrinsic control.
\newblock \emph{Workshop at the International Conference on Learning Representations}, 2017.

\bibitem[Hansen et~al.(2020)Hansen, Dabney, Barreto, Warde-Farley, de~Wiele, and Mnih]{hansen2020Fast}
Steven Hansen, Will Dabney, Andre Barreto, David Warde-Farley, Tom~Van de~Wiele, and Volodymyr Mnih.
\newblock Fast task inference with variational intrinsic successor features.
\newblock In \emph{International Conference on Learning Representations}, 2020.

\bibitem[Hu et~al.(2024)Hu, Wang, Stone, and Mart\'{\i}n-Mart\'{\i}n]{hu2024dusdi}
Jiaheng Hu, Zizhao Wang, Peter Stone, and Roberto Mart\'{\i}n-Mart\'{\i}n.
\newblock Disentangled unsupervised skill discovery for efficient hierarchical reinforcement learning.
\newblock In \emph{Advances in Neural Information Processing Systems}, volume~37, pp.\  76529--76552, 2024.

\bibitem[Kim et~al.(2021)Kim, Park, and Kim]{kim2021ibol}
Jaekyeom Kim, Seohong Park, and Gunhee Kim.
\newblock Unsupervised skill discovery with bottleneck option learning.
\newblock In \emph{Proceedings of the 38th International Conference on Machine Learning}, volume 139, pp.\  5572--5582, 2021.

\bibitem[Krakovna et~al.(2020)Krakovna, Orseau, Ngo, Martic, and Legg]{krakovna2020avoiding}
Victoria Krakovna, Laurent Orseau, Richard Ngo, Miljan Martic, and Shane Legg.
\newblock Avoiding side effects by considering future tasks.
\newblock \emph{Advances in Neural Information Processing Systems}, 33:\penalty0 19064--19074, 2020.

\bibitem[Lee et~al.(2020)Lee, Yang, and Lim]{lee2020Learning}
Youngwoon Lee, Jingyun Yang, and Joseph~J. Lim.
\newblock Learning to coordinate manipulation skills via skill behavior diversification.
\newblock In \emph{International Conference on Learning Representations}, 2020.

\bibitem[Liu \& Abbeel(2021)Liu and Abbeel]{liu2021aps}
Hao Liu and Pieter Abbeel.
\newblock Aps: Active pretraining with successor features.
\newblock In \emph{Proceedings of the 38th International Conference on Machine Learning}, volume 139, pp.\  6736--6747, 2021.

\bibitem[Miyato et~al.(2018)Miyato, Kataoka, Koyama, and Yoshida]{miyato2018spectral}
Takeru Miyato, Toshiki Kataoka, Masanori Koyama, and Yuichi Yoshida.
\newblock Spectral normalization for generative adversarial networks.
\newblock \emph{arXiv preprint arXiv:1802.05957}, 2018.

\bibitem[Park et~al.(2022)Park, Choi, Kim, Lee, and Kim]{park2022lsd}
Seohong Park, Jongwook Choi, Jaekyeom Kim, Honglak Lee, and Gunhee Kim.
\newblock Lipschitz-constrained unsupervised skill discovery.
\newblock In \emph{International Conference on Learning Representations}, 2022.

\bibitem[Park et~al.(2023)Park, Lee, Lee, and Abbeel]{park2023controllability}
Seohong Park, Kimin Lee, Youngwoon Lee, and Pieter Abbeel.
\newblock Controllability-aware unsupervised skill discovery.
\newblock In \emph{Proceedings of the 40th International Conference on Machine Learning}, volume 202, pp.\  27225--27245, 2023.

\bibitem[Park et~al.(2024)Park, Rybkin, and Levine]{park2024metra}
Seohong Park, Oleh Rybkin, and Sergey Levine.
\newblock {METRA}: Scalable unsupervised {RL} with metric-aware abstraction.
\newblock In \emph{The Twelfth International Conference on Learning Representations}, 2024.

\bibitem[Peng et~al.(2022)Peng, Hu, Zhang, Tang, Guo, Yi, Chen, zhang, Du, Li, Guo, and Chen]{hu2022causality}
Shaohui Peng, Xing Hu, Rui Zhang, Ke~Tang, Jiaming Guo, Qi~Yi, Ruizhi Chen, xishan zhang, Zidong Du, Ling Li, Qi~Guo, and Yunji Chen.
\newblock Causality-driven hierarchical structure discovery for reinforcement learning.
\newblock \emph{Advances in Neural Information Processing Systems}, 35:\penalty0 20064--20076, 2022.

\bibitem[Sharma et~al.(2020)Sharma, Gu, Levine, Kumar, and Hausman]{sharma2020dynamics}
Archit Sharma, Shixiang Gu, Sergey Levine, Vikash Kumar, and Karol Hausman.
\newblock Dynamics-aware unsupervised discovery of skills.
\newblock In \emph{International Conference on Learning Representations}, 2020.

\bibitem[Sutton et~al.(1999)Sutton, Precup, and Singh]{SUTTON1999181}
Richard~S. Sutton, Doina Precup, and Satinder Singh.
\newblock Between mdps and semi-mdps: a framework for temporal abstraction in reinforcement learning.
\newblock \emph{Artificial Intelligence}, 112\penalty0 (1):\penalty0 181--211, 1999.

\bibitem[Turner et~al.(2020)Turner, Hadfield-Menell, and Tadepalli]{turner2020conservative}
Alexander~Matt Turner, Dylan Hadfield-Menell, and Prasad Tadepalli.
\newblock Conservative agency via attainable utility preservation.
\newblock In \emph{Proceedings of the AAAI/ACM Conference on AI, Ethics, and Society}, pp.\  385--391, 2020.

\bibitem[Wang et~al.(2024)Wang, Hu, Chuck, Chen, Mart{\'\i}n-Mart{\'\i}n, Zhang, Niekum, and Stone]{wang2024skild}
Zizhao Wang, Jiaheng Hu, Caleb Chuck, Stephen Chen, Roberto Mart{\'\i}n-Mart{\'\i}n, Amy Zhang, Scott Niekum, and Peter Stone.
\newblock Skild: unsupervised skill discovery guided by factor interactions.
\newblock In \emph{Advances in Neural Information Processing Systems}, volume~37, pp.\  77748--77776, 2024.

\bibitem[Zhang et~al.(2021)Zhang, Yu, and Xu]{zhang2021hierarchical}
Jesse Zhang, Haonan Yu, and Wei Xu.
\newblock Hierarchical reinforcement learning by discovering intrinsic options.
\newblock In \emph{International Conference on Learning Representations}, 2021.

\end{thebibliography}
\bibliographystyle{rlj}

\appendix
\section{Algorithms for Focused DIAYN and LSD}
\label{appendix:focused-algorithms}
In this section, we describe the additional algorithms used in our experiments.

\begin{algorithm}[H]
\caption{Focused Diversity is All You Need}\label{alg:fdiayn}
\begin{algorithmic}
\For{episode $=1,M$}
\State Sample $s_0$ from the initial state distribution $\mu$
\State Sample skill $z$ from $\nu$
\For{$t=1,t_{\text{max}}$}
\State Select action $a_t\sim \pi_z(\cdot |s_t)$
\State Observe $s_{t+1}\sim p(\cdot |s_t, a_t)$
\State Calculate the skill reward $r_{\text{focused-DIAYN}}(s_0,z,s_{t+1})$ using Equation~\ref{eq:focused-diayn}.
\State Update policy $\pi_z$
\For{$i\in \mathcal{V}_z$}
\State Update the skill discriminators $d_{i}$ from $(z^i, s_{t+1}^{i})$ 
\EndFor
\EndFor
\EndFor
\end{algorithmic}
\end{algorithm}

\begin{algorithm}[H]
\caption{Focused Lipschitz-constrained Skill Discovery}\label{alg:flsd}
\begin{algorithmic}
\For{episode $=1,M$}
\State Sample $s_0$ from the initial state distribution $\mu$
\State Sample skill $z$ from $\nu$
\State Follow policy $\pi_z$ until termination state $s_T$
\For{$i\in \mathcal{V}_z$}
\State Update $\phi_i$ using $(s_0^i, z^i, s_T^{i})$ 
\EndFor
\State Calculate the reward $r_{\text{focused-LSD}}(s_0, z, s_{T})$ using Equation~\ref{eq:focused-lsd}.
\State Update $\pi_z$ to maximize $r_{\text{focused-LSD}}$
\EndFor
\end{algorithmic}
\end{algorithm}
\section{Additional Training Details}
\label{appendix:hyperparameters}
Following prior work~\citep{gregor2017vic,eysenbach2018diayn,park2022lsd}, the skill distribution $\nu$ was uniform over all skills and held constant during training. The skill discriminators for all mutual-information-based skill discovery algorithms made predictions using an exponentially weighted moving average of the previous samples.

For LSD, the state representation function $\phi$ was a linear function of the state trained with stochastic gradient descent. Following~\citeauthor{park2022lsd}, we used spectral normalization~\citep{miyato2018spectral} to satisfy the Lipschitz constraint during training.

\subsection{Hyperparameters}
The skill discovery phase had hyperparameters for both skill policies and skill rewards. All algorithms used the same skill policy hyperparameters but different skill reward hyperparameters, as VIC, DIAYN, and LSD generate rewards differently.

For skill policies, we searched for the discount factor $\gamma$ in $\{0.9, 0.99, 0.999\}$, Q-table learning rate $\alpha$ in $\{0.1, 0.01\}$, and exponential learning rate decay parameter $\kappa$ in $\{0.001, 0.0005, 0.0001, 0.0002, 0.0001\}$. We chose $\gamma=0.99, \alpha=0.1, \kappa=0.0005$ by finding the values that maximized the baseline VIC performance in FourRooms. Initial values in the Q-table were set to 0. We chose the maximum number of steps per skill (40 in FourRooms, 20 in ForageWorld/MudWorld) by calculating the max number of steps needed for a focused skill to terminate and rounding up to the next multiple of 10. A larger number of steps did not seem to affect performance, as skills typically chose to terminate early.

For skill rewards, we searched for the exponentially weighted moving average (EMWA) weight of the VIC skill discriminator in $\{0.3, 0.5, 0.7\}$, the EMWA weight of DIAYN skill discriminator in $\{0.001, 0.05, 0.01\}$, and the stochastic gradient descent (SGD) learning rate of the LSD state representation function $\phi$ in $\{0.1, 0.01\}$. We chose the skill reward hyperparameters by finding the value that led to the best performance of the baseline skill discovery algorithm and using it for both the DUSDi and focused versions of the algorithm. All VIC algorithms (baseline, DUSDi, and focused) used an EWMA weight of 0.7 in FourRooms, 0.5 in ForageWorld and 0.7 in MudWorld. All DIAYN algorithms (baseline, DUSDi, and focused) used an EWMA weight of 0.05 in all environments. All LSD algorithms (baseline and focused) used an SGD learning rate of 0.1 in all environments. Following~\citet{hu2024dusdi}, we used the same hyperparameters and skill discriminator architecture for the skill discriminator of the DUSDi side effects penalty.

We tried penalty strengths in $\{2, 4, 10, 40\}$ for focused LSD/DIAYN, $\{1, 2, 8, 10\}$ for focused LSD and $\{0.01, 0.1, 0.2, 0.5\}$ for DUSDi VIC/DIAYN. We chose the penalty strength by using the value that led to the best skill learning performance in FourRooms. The final values were: 4 for focused VIC/DIAYN, 2 for focused LSD, and 0.1 for DUSDi VIC/DIAYN. Re-tuning the DUSDi penalties in ForageWorld and MudWorld did not yield performance gains (this is consistent with Section~\ref{subsec:dusdi-comparison}).

We fixed the number of skills at 16 but also ran the FourRooms task with 8 skills (8 tool skills, 0 navigation skills) and 24 skills (8 tool skills, 16 navigation skills). The results were qualitatively similar to Figure 3 (left): baseline skills didn’t solve the task, while the DUSDi and focused skills showed comparable performance.

In downstream tasks, we searched for the exploration rate decay parameter of the skill-selection policy in $\{0.0002, 0.0005, 0.001, 0.002\}$, keeping its other hyperparameters the same as the skill policies (a discount factor of 0.99, a Q-table learning rate of 0.1 and Q-table initialized to 0). Because the baseline algorithms did not converge, we selected decay rates per algorithm. In both FourRooms and MudWorld, all algorithms used a decay rate of 0.001. In ForageWorld, skill-selection policies with focused skills used a decay rate of 0.001 while the remaining policies used a decay rate of 0.0005. Hyperparameters weren't changed for underspecified tasks (Section~\ref{subsec:underspec}).


\end{document}